\documentclass[conference]{IEEEtran}
\usepackage{cite}
\usepackage{amsmath,amssymb,amsfonts}
\usepackage{graphicx}
\usepackage{booktabs}
\usepackage{iftex}

\ifPDFTeX
  \usepackage[utf8]{inputenc}
\fi

\def\BibTeX{{\rm B\kern-.05em{\sc i\kern-.025em b}\kern-.08em
    T\kern-.1667em\lower.7ex\hbox{E}\kern-.125emX}}

\begin{document}

\title{DeVIT: Low-Power Vision Transformer Acceleration Using Delta Computation}

\author{%
  \IEEEauthorblockN{Reyhaneh Hosseinzadeh, Parham Zilouchian Moghaddam, Mehdi Modarressi}
  \IEEEauthorblockA{%
    \textit{School of Electrical and Computer Engineering, University College of Engineering, University of Tehran, Tehran, Iran}\\
    Email: \{hossseinzadeh.ray, p.zilouchian, modarressi\}@ut.ac.ir
  }
}

\maketitle

\begin{abstract}

The emergence of transformer-based deep learning models has brought unprecedented performance across various domains, particularly in natural language processing and computer vision. However, deploying these models, especially on resource-constrained devices, poses significant challenges due to their high computational complexity and large memory size and bandwidth requirements. This complexity has led researchers to use low-bit model weights to reduce memory usage and improve efficiency. In addition to reducing processing and memory demands, quantization introduces another useful property: \emph{value locality}, where the extremely large number of parameters are restricted to a limited range of values. To fully take advantage of this locality, this paper presents DeVIT, an acceleration method for vision transformers that leverages differential computation to enable multiplier-less matrix multiplication.

By storing weight matrices in a differential format and calculating each inner partial product of matrix multiplication by reusing results from previous computations, the proposed approach simplifies matrix multiplication, an essential operation in transformer layers, into basic shift-and-add processes. DeVIT reduces the computational load of Vision Transformers beyond what is achievable with quantization alone and enhances the practicality of deploying transformer models for image processing tasks across a wider range of devices. In the evaluated configurations, DeVIT reduces normalized computation load to $0.53$ of an unoptimized baseline. Its single-GEMM energy is $159.57$\,nJ, approximately $5.5\%$ below the lowest-energy ShiftAddLLM configuration included in our comparison.

\end{abstract}


\section{Introduction}

Vision Transformers (ViTs) have recently shown superior performance in image processing by leveraging self-attention to model long-range dependencies effectively. However, these advancements come with substantial computational and memory requirements. Unlike Convolutional Neural Networks (CNNs), ViTs exhibit quadratic complexity in sequence length for self-attention, resulting in a significant increase in computational load, primarily due to multiply-accumulate (MAC) operations~\cite{dosovitskiy2020image,carion2020end,Swin,VITxue2025similarity,pan2025stream}.

For example, the ViT-B/16 model contains 86 million parameters and requires 17.6 billion MAC operations for processing each $224\times224$ image, in contrast to (the already high) 4.1 billion MAC operations for ResNet-50. Larger models, such as ViT-L/16---with 307 million parameters and requiring 61 billion MACs, push the boundaries of computational load even further.
This huge computation load renders real-time and energy-efficient deployment of ViTs almost impractical without software and hardware optimization/customization methods.
Quantization has long been the primary mechanism to reduce the complexity of neural networks~\cite{daneshtalab2020hardware}.
Recent research shows that ViTs are highly amenable to quantization and can operate effectively with low-bit parameters~\cite{ptq4vit,iviT,li2022q}. In low-bit quantization (such as 8-bit), each weight is mapped to one of 256 discrete values. Given the scale of ViT weight matrices, it is highly likely that nearly all of these quantized values appear within the matrix. This characteristic, \emph{locality of values}, is a secondary benefit of quantization and presents opportunities for simplifying multiplication.

For instance, consider a row from the input embedding matrix: each element in this row is multiplied by all elements in the respective row of the three projection weight matrices (for queries, keys, and values) to produce partial sums for the corresponding output cells. Since the row of weights is likely to encompass most of the unique values within the quantization range, the same input element will eventually be multiplied by all possible quantized weight values.

To leverage this distinctive property to reduce the computational complexity of ViTs, this paper introduces DeVIT, a \emph{De}lta-coded \emph{Vi}sion \emph{T}ransformer. DeVIT sorts the weights within each row of the weight matrix in ascending order and stores them in a differential format (delta format). This format represents the differences between consecutive weights in the sorted row, replacing traditional multiplication with differential computation and enabling the systematic reuse of partial results for enhanced hardware efficiency.

Computations begin by multiplying the input element by the smallest weight in the row, which is now located at the head of the sorted row. For the subsequent weights, results are generated by reusing the previous product and applying only the difference between the current and prior weights. Thanks to the value locality introduced by low-bit quantization, these differences are typically minor (mostly ranging between 0 and 2, as will be demonstrated later) and can be efficiently approximated using the nearest power of two. This allows for the substitution of multiplication of the difference and the input element with a simple shift operation. Further, low-bit quantization ends up having many identical values in each row, represented by zero delta: This opportunity can be exploited by skipping the computation of zero delta values and reusing the result for all subsequent weights as long as the delta remains zero.

This differential approach markedly reduces the number of expensive multiplications, significantly improving the hardware efficiency of ViTs. In this paper, we first describe how the weight matrices and associated computations are reorganized to enable delta computations. We then present the DeVIT architecture, which efficiently implements the proposed delta-based method. In particular, we show how the architecture reorders the weight matrices and retrieves the correct weight positions after the weights have been reordered for differential computation.

The concept of computation reuse by exploiting value locality in weights~\cite{mahdiani2019delta,ghanbari2022energy,yasoubi2016power,ucnn,khodarahmi2024remove,skippynn} and input data~\cite{jose18,flich2025sirena,deltarnn,modarressi2016low} has been explored in prior work for earlier neural architectures, such as feedforward, convolutional, and recurrent neural networks. In our previous work $\Delta$NN~\cite{mahdiani2019delta}, we showed that applying differential computing can considerably reduce the power usage of convolutional neural networks. Transformers require a different organization: their weights form large dense projection and feed-forward matrices rather than small spatial filters; their activations vary by token rather than being reused as fixed spatial neighborhoods; and their attention path interleaves dense projections with softmax-weighted reductions. DeVIT therefore introduces a transformer-specific dataflow, partitioning strategy, and index-management scheme.

The experimental results show a normalized computation load of $0.53$ relative to the unoptimized baseline and a measured single-GEMM energy of $159.57$\,nJ. Detailed accuracy, computation-load, and energy results are presented in Section~\ref{sec:experimental}.

\section{Preliminary and Motivation}

\label{sec:motivation}

\textbf{Vision Transformers in vision.}
ViTs adapt the transformer approach from NLP to images by treating an image as a sequence of patches. This tokenization allows for long-range interactions to become fundamental, resulting in models that are architecturally simple, scalable, and effective in recognition tasks.

\textbf{Why ViTs are an important accelerator target.}
Unlike a CNN, a ViT can model interactions between distant image regions in every encoder layer. This flexibility has enabled a common family of transformer backbones to support classification, detection, and dense prediction~\cite{dosovitskiy2020image,touvron2021training,carion2020end,Swin}. The benefit comes with substantial dense-projection and feed-forward computation, making those operations a practical target for hardware acceleration.

At a high level, a standard ViT is primarily encoder-centric. The encoder is composed of stacked layers that implement Multi-Head Self-Attention (MHSA) and an FFN, which progressively refine the representations of input tokens. Although traditional ViTs do not include a decoder, many downstream tasks, such as segmentation and image generation, include decoder heads that leverage encoder features to generate structured output. In this work, we focus on the encoder, as it plays a dominant role in both computational and memory requirements.

\textbf{Layer anatomy.}
Each encoder layer comprises three parts: (i) \emph{Linear Projections} that form queries, keys, and values (QKV) and an \emph{output projection}; (ii) \emph{MHSA}, where dot products between Q and K are softmax-normalized and applied to V; and (iii) an \emph{FFN} implemented as an expansion MLP followed by a contraction back to the hidden size. Three hyperparameters shape a layer: hidden size, FFN size, and number of heads. ViTs often surpass CNNs in accuracy, but with larger parameter counts and higher per-layer compute.

\begin{figure}[!t]
  \centering
  \includegraphics[width=\linewidth]{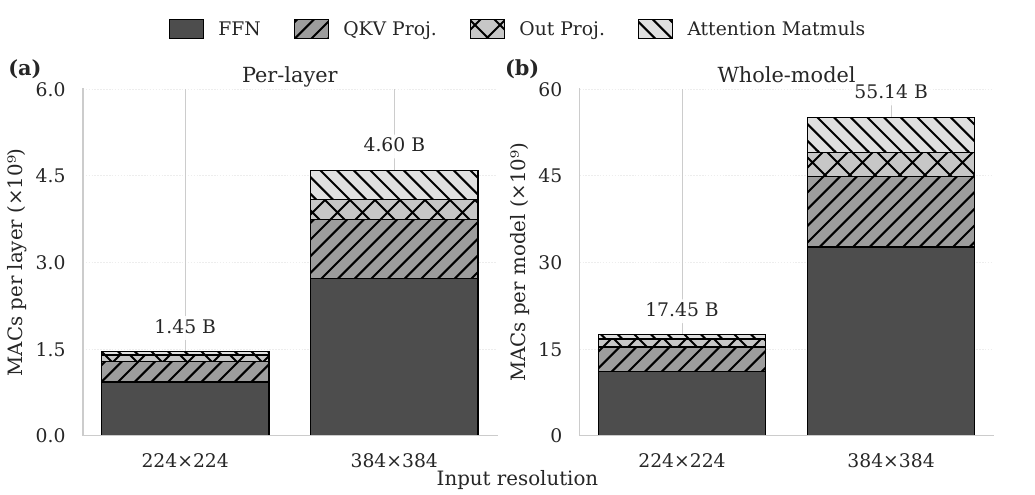}
  \caption{MAC breakdown of ViT-B/16 at two input resolutions:
  (a)~per encoder layer and (b)~whole model. Dense linear layers (FFN and
  projections) dominate; attention matmuls are comparatively minor. Softmax and
  LayerNorm are excluded (${<}1\%$ of per-layer FLOPs).}
  \label{fig:vit_compute}
\end{figure}

\textbf{Where ViTs spend compute, and why it matters.}
Fig.~\ref{fig:vit_compute} separates the encoder's work into QKV projections, the output projection, attention matrix multiplies, and the FFN. It shows that GEMM-based dense linear layers, which are QKV projections and the FFN, dominate the computation. This observation is consistent with recent ViT acceleration work targeting short-token regimes, where these parts are reported to be the dominant bottleneck~\cite{hsiung2025lpvit}. Even as sequence length increases and attention cost grows, GEMMs remain the primary bottleneck. Our delta-based method reduces GEMM complexity within these layers, enabling effective acceleration for GEMM-dominated ViTs while preserving the original model structure.

\textbf{Scope and assumptions.}
Unless otherwise specified, we analyze encoder-only Vision Transformers (ViTs) defined by a hidden size of $d$, $H$ attention heads, and a feed-forward network (FFN) expansion given by $d_{\text{ff}} = 4d$. The sequence length $N$ is calculated based on non-overlapping $16 \times 16$ patches plus a class token; for example, $N = 197$ for $224^2$ input images and $N = 577$ for $384^2$ input images. We adopt the convention $Y = X W$, where $X \in \mathbb{R}^{N \times d}$ is the input matrix, $W \in \mathbb{R}^{d \times d_{\text{out}}}$ is the weight matrix, and each output element $Y[i,j] = \sum_{k} X[i,k]\, W[k,j]$.

\section{Related Work}

Algorithmic approaches to efficient ViTs seek a balance between computational cost and accuracy. These approaches can be divided into four main categories:
\begin{enumerate}
  \item \textbf{Memory-aware methods}: These techniques focus on minimizing data movement within the attention and feed-forward layers, for example via tensor decomposition or low-rank factorization of the projection matrices.
  \item \textbf{Computation-aware methods}: This category aims to lower the arithmetic cost by applying techniques such as quantization or promoting sparsity. Notable examples are Q-ViT~\cite{li2022q}, PTQ4ViT~\cite{ptq4vit}, and I-ViT~\cite{iviT}, which together cover quantization-aware training, post-training quantization, and integer-only inference of ViTs.
  \item \textbf{Architecture-aware methods}: These methods restructure the model to enhance efficiency. This can involve the introduction of hierarchical attention or the merging of patches, as demonstrated by the Swin Transformer~\cite{Swin} and RePaViT~\cite{xu2025repavit}.
  \item \textbf{Token-level methods}: Orthogonal to GEMM-side optimizations, dynamic token pruning and merging---such as DynamicViT~\cite{dynamicvit} and Token Merging (ToMe)~\cite{tome}---reduce sequence length conditionally on input content, complementing rather than competing with weight-side techniques like ours.
\end{enumerate}

On the other hand, \textbf{system-aware methods} extend beyond purely algorithmic optimizations. They combine algorithmic modifications---related to data, parameters, or dataflow---with corresponding hardware support. This approach facilitates the efficient execution of the modified model.

System-aware accelerators for ViTs can be categorized into three broad groups:
\begin{enumerate}
  \item The first group focuses on minimizing computational workloads by sparsifying the attention matrix, often through precomputation or predictive algorithms that identify key attention patterns. A representative example is ViTCoD~\cite{ViTCoD}, which utilizes a learned mask to differentiate between dense and sparse workloads across specialized hardware. HeatViT~\cite{heatvit} similarly leverages adaptive token pruning combined with 8-bit quantization on embedded FPGAs. However, it is important to note that, in ViTs, the primary computational cost does not arise solely from the attention mechanism.
  \item The second category addresses the overhead associated with linear transformations. Methods in this group include weight sparsification, such as structured M:N sparsity, which prunes weight matrices during training in an architecture-aware manner~\cite{bambhaniya2024progressive,fang2022algorithm}. Additionally, some approaches utilize lookup tables (LUTs) to minimize redundant computations. Examples of this include ShiftAddViT for Vision Transformers, ShiftAddLLM for large language models, LUT Tensor Core, and the earlier ShiftAddNet design that originated this family~\cite{you2024shiftaddvit,you2024shiftaddllm,mo2025lut,shiftaddnet}. While these methods utilize low-bit weights through partial-product reuse, they introduce additional overhead from LUT access, require extra storage for repetitive activations, and still depend on dynamic activations, necessitating online computation.
  \item The third category seeks to simultaneously reduce computations in both linear projections and attention mechanisms. For instance, AccelTran~\cite{acceltran} leverages weight sparsity identified during pretraining or fine-tuning and applies dynamic activation sparsity through a learned threshold on the attention matrix. However, its reductions primarily target attention, which is not the primary cost in ViTs. Similarly, FACT~\cite{fact} optimizes both attention and the generation of Query, Key, and Value (QKV) components and reduces the costs associated with Feed-Forward Networks (FFN) through token quantization. SwiftTron~\cite{swifttron} pushes this further with an integer-only Transformer ASIC handling non-linear operations directly in fixed-point arithmetic. Nonetheless, these methods operate on activations and, as a result, incur considerable online precomputation overhead.
\end{enumerate}

Previous works on CNNs, such as $\Delta$NN, C-CORN, and SkippyNN~\cite{mahdiani2019delta,ghanbari2022energy,skippynn}, have aimed to improve efficiency through computational reuse, sorting weights and applying a differential MAC (DMAC) unit to share products across a row. Power-of-two and shift-based weight representations~\cite{apot} provide a complementary line of work that turns multiplications into shifts at the cost of a small quantization error. Our approach for transformers similarly capitalizes on computation reuse to eliminate redundant operations across the network but adapts the dataflow and storage scheme to the dense-GEMM structure characteristic of ViTs.

\section{Differential Computing for ViTs}
\label{sec:diffcomp}

\textbf{Differential computation opportunity.} Our architecture enhances the efficiency of matrix multiplication in scenarios where one matrix---the weight matrix---is fixed and can be processed offline during the design phase. With low-bit quantization, all potential quantized values will likely manifest across the matrix and even within individual rows, so there are many weights with repetitive and close values. However, during matrix multiplication, the output cells are computed sequentially, and each weight is multiplied by a different input element. Because of this, reusing the result of a multiplication with weight value $v$ for another weight with value $v{+}1$ is generally not feasible.

Reuse becomes possible only when two weights are multiplied by the same input value, allowing the multiplication result to be shared and reused. Under the convention $Y = X W$ (introduced in Section~\ref{sec:motivation}), the $k$-th element of an input row $X[i,k]$ is multiplied by every entry of the $k$-th \emph{row} of $W$---that is, $W[k,:]$---and the resulting partial products are scattered across the $i$-th row of the output, one per output column $j$. In other words, the input element $X[i,k]$ contributes to every output element $Y[i,:]$ through the weights in row $k$ of $W$. This common input operand is exactly what makes computation reuse possible: techniques such as delta coding and result reuse can be applied across the sequence of weights associated with each input element, reducing redundant computation.

\noindent By adopting an \emph{input-stationary} execution order, where input elements are fetched sequentially and reused across all required computations, many multiplications share the same input operand. This creates a scenario where one operand remains constant across multiple multiplications and will eventually be multiplied by every quantized value, albeit in varying sequences. Consequently, the weights in each row of $W$ are arranged in ascending order to promote reusing the multiplication result of each weight for subsequent weights.

\begin{figure}[!t]
    \centering
    \includegraphics[trim=3 3 3 3,clip,width=\columnwidth]{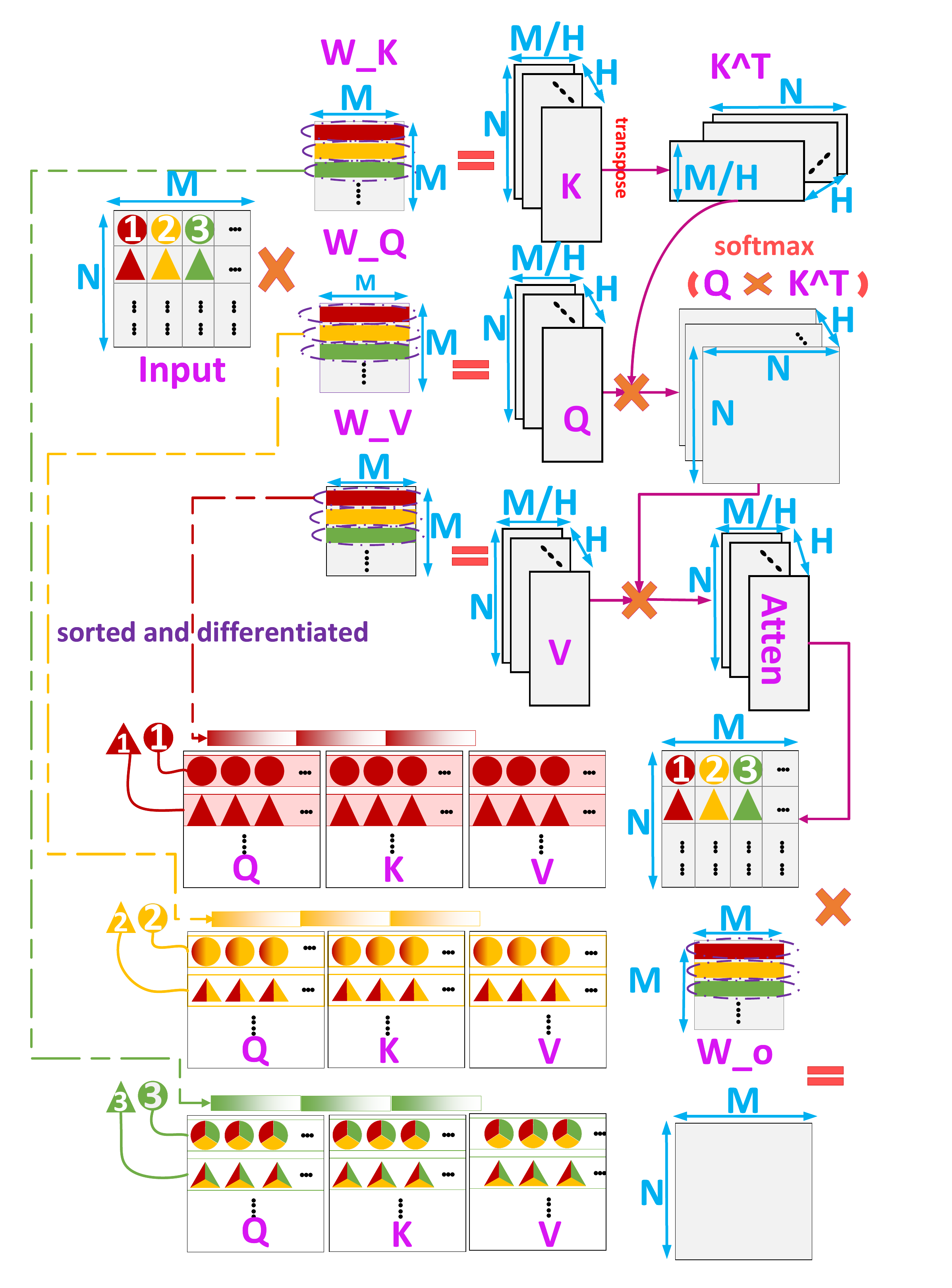}
    \caption{The general architecture of multi-head self-attention and input-stationary data flow of linear generation with delta format.}
    \label{fig:model_arch}
\end{figure}

\begin{figure*}[!t]
    \centering
    \includegraphics[width=\textwidth]{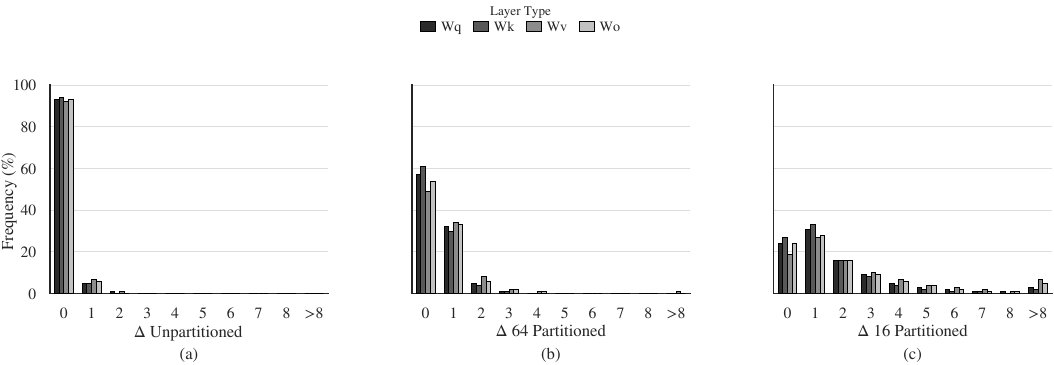}
    \caption{Delta distribution of the QKV and output projection weights
    ($W_q$, $W_k$, $W_v$, $W_o$) of ViT-B/16, averaged over all 12 encoder
    layers: (a)~unpartitioned, (b)~64-element partitions, and
    (c)~16-element partitions.}
    \label{fig:delta-distribution-partitioning}
\end{figure*}

\begin{figure*}[!t]
    \centering
    \includegraphics[width=\textwidth]{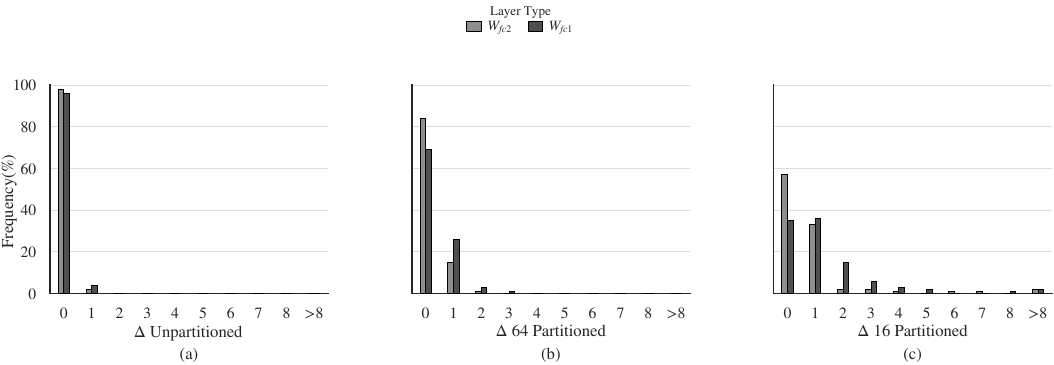}
    \caption{Delta distribution of the FFN weights ($W_{\text{fc1}}$,
    $W_{\text{fc2}}$) of ViT-B/16, averaged over all 12 encoder layers:
    (a)~unpartitioned, (b)~64-element partitions, and
    (c)~16-element partitions.}
    \label{fig:delta-distribution-FC}
\end{figure*}

\textbf{The dataflow of DeVIT.} The overall data flow and computation reuse strategy are illustrated in Fig.~\ref{fig:model_arch}.\noindent Given a sorted row of $W$, we compute the product of the input $a$ with the next weight $W_2$ as follows:

\begin{equation}
  a \times W_2 \;=\; a \times W_1 \;+\; a \times \Delta W,
  \label{eq:incremental-product}
\end{equation}

where $W_1$ is the preceding weight in the sorted order and $\Delta W = W_2 - W_1$.

\noindent Given the large size of each matrix row, many quantized values repeat or lie close together. For example, an 8-bit, hidden-size-$768$ row contains $768$ elements drawn from only $256$ possible values. Sorting such a row therefore produces many zero or small deltas. The measured distributions in Figs.~\ref{fig:delta-distribution-partitioning} and~\ref{fig:delta-distribution-FC} confirm that zero and one dominate in the unpartitioned case. Zero deltas reuse the preceding product, while power-of-two deltas replace multiplication with a shift and an addition.

\noindent Rather than storing absolute weight values, DeVIT employs delta values. In this system, during the delta operation---which replaces multiplication---the weight is interpreted as a delta value, the input is shifted accordingly, and the result is added to the output of the preceding multiplication.

\noindent Storing delta values changes the order of weights within each row. To restore the original output position, each delta value is associated with its corresponding index $j$, indicating the original column of the weight in the row. This index determines the output cell $(i,j)$ into which the partial sum is accumulated. The total storage per weight is therefore the sum of the delta-code width and the within-partition index width; for a 64-element partition, this amounts to $10$ bits per weight. Section~\ref{sec:index} describes the encoding in detail.

\noindent For a quantitative evaluation, we present the \emph{Delta Distribution} of the weights $W_q$, $W_k$, $W_v$, and $W_o$ for the ViT-B/16 model in Section~\ref{sec:experimental}. These weights have been quantized to $8$ bits, sorted, and differentiated to facilitate a thorough analysis of their distribution.

\noindent The ViT-B/16 model has a hidden size of $768$, an MLP size of $3072$, and $12$ attention heads. The ``16'' denotes the side length of each input patch. The individual projection matrices $W_q$, $W_k$, $W_v$, and $W_o$ have dimensions $M \times M$; an implementation that concatenates the first three represents $W_{qkv}$ as $M \times 3M$. Each row is sorted before the differences between consecutive weights are encoded. The distributions are averaged first across rows and then across all $12$ encoder layers.

\noindent Partitioning trades some of this concentration for a smaller output buffer. As the partition size decreases from the unpartitioned case to $64$ and then $16$ elements, the distributions spread toward larger deltas, but zero and small deltas remain common (Figs.~\ref{fig:delta-distribution-partitioning} and~\ref{fig:delta-distribution-FC}).

\section{DeVIT Architecture}

\begin{figure}[!t]
    \centering
    \includegraphics[trim=3 3 3 3,clip,width=\columnwidth]{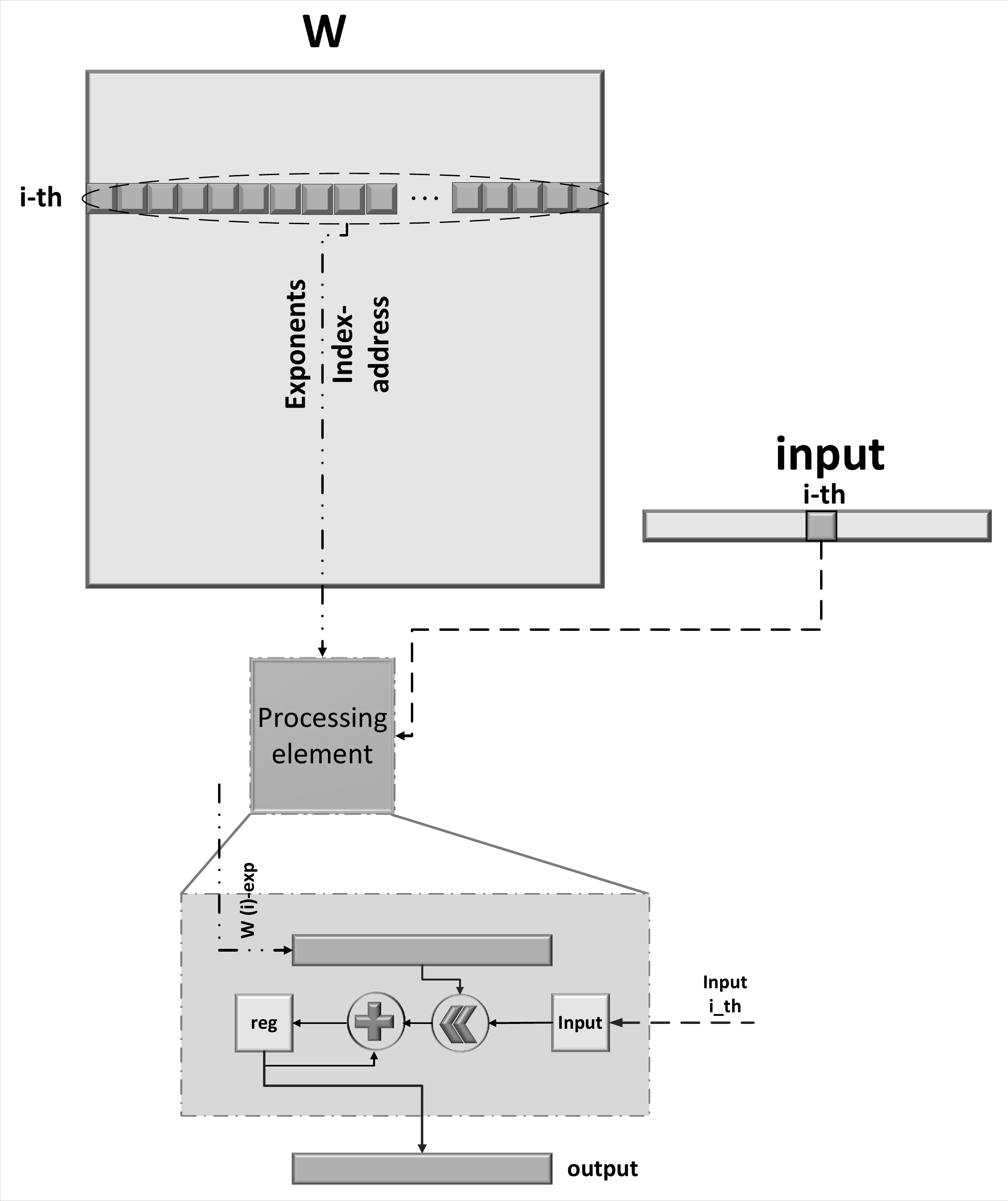}
    \caption{The unpartitioned DeVIT core: delta-encoded weights and their index addresses drive a shift--add processing element that accumulates partial sums into the output buffer.}
    \label{fig:Delta_vit_core}
\end{figure}

Figure~\ref{fig:Delta_vit_core} shows the unpartitioned DeVIT processing element. A weight buffer stores the encoded differences $\Delta W$ and their destination indices. Each nonzero magnitude is approximated by a supported power of two, allowing the processing element to use shifts and additions while an output buffer accumulates the partial sums.
For an activation $a_{ik}$, the processing element traverses row $k$ of $\Delta W$. The current product is retained in the \texttt{reg} register and updated by each subsequent shift--add operation. The associated destination index routes each partial product to the correct output element. For an $N \times M$ input, an $M \times M$ weight matrix, and $K$ weight entries processed in parallel, the cycles required per input row are
\[\left\lceil \frac{M}{K} \right\rceil \times M.\]
The unpartitioned DeVIT approach involves a high number of on-the-fly partial products, leading to increased power consumption. To overcome the challenges related to buffer size and power usage in neural network computations, a strategic weight matrix management approach is utilized. The buffer used for storing partial products can be significant, contributing to higher power consumption, and its size is directly linked to the number of columns in the weight matrix.

To address these issues, the weight matrices are partitioned into appropriate sizes before applying the delta method and input-stationary approach to each resulting block of weights. This partitioning strategy offers several advantages: it reduces the required buffer size as each partitioned block is smaller than the original matrix; potentially decreases power consumption due to the reduced buffer size; allows for more efficient application of the delta method and input-stationary approach to these smaller blocks; and enables more flexible processing, potentially facilitating parallel computation of different blocks.

By implementing this partitioning technique along with delta encoding and input-stationary dataflow, the overall efficiency of neural network computations can be further improved, achieving a balance between performance and resource utilization.

We analyzed the ViT-B/16 weights in blocks and measured the delta distribution across all $12$ layers. Frequencies were averaged across rows within each block, across blocks, and finally across layers. Figure~\ref{fig:delta-distribution-partitioning} shows the expected trend: 64-element partitions retain a strong concentration at small deltas, whereas 16-element partitions spread more probability toward larger magnitudes and therefore offer less computation reuse.

\begin{figure}[!t]
    \centering
    \includegraphics[trim=3 3 3 3,clip,width=\columnwidth]{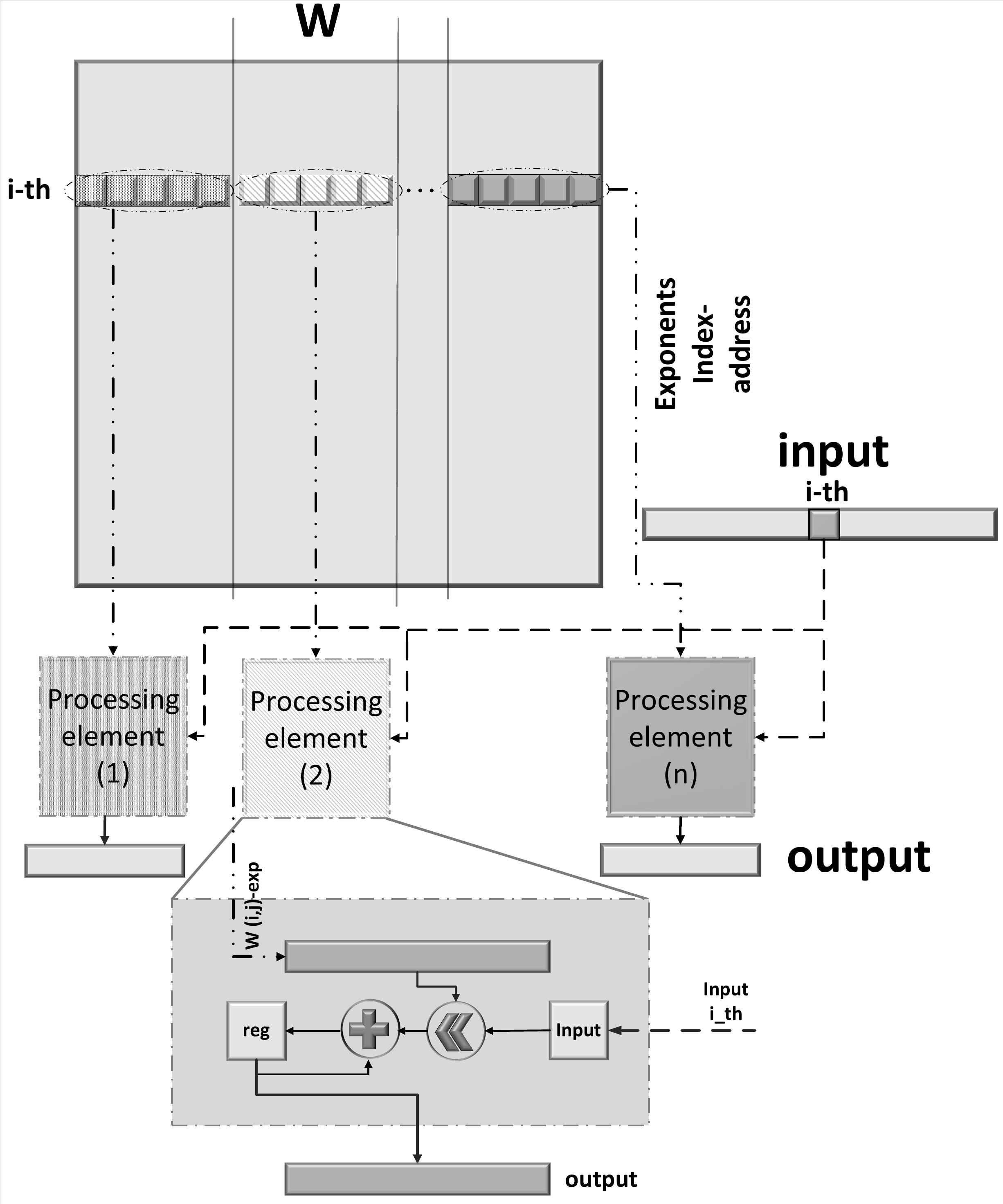}
    \caption{Partitioned DeVIT architecture for producing and routing partial products from delta-encoded weights.}
    \label{fig:Delta_vit_core_partitioned}
\end{figure}

In the diagram presented in Fig.~\ref{fig:Delta_vit_core_partitioned}, we initially partition the weights, followed by sorting and encoding them in the delta format. We apply the Delta method and input stationary to generate partial products, which are then stored in an output buffer. Subsequently, the elements of each column are restored to their correct positions and accumulated. Through the partitioning process, we achieve both input-stationary and output-stationary mechanisms, as the first input must be read for the subsequent partitions.

For a given input matrix, weight matrix, batch size of $K$, and partition size $S$, the number of cycles can be calculated using the following formula:
\[ \left\lceil \frac{M}{K} \right\rceil \times S. \]

\section{Index Encoding}
\label{sec:index}

\noindent Each delta uses a 4-bit code comprising one sign bit and a 3-bit magnitude field. The magnitude field reserves one code for zero and represents the supported power-of-two magnitudes $\{1,2,4,8\}$; unused code points are reserved. Deltas beyond the supported range (the ``$>\!8$'' category in Figs.~\ref{fig:delta-distribution-partitioning} and~\ref{fig:delta-distribution-FC}) are saturated at the largest representable magnitude. This approximation introduces the accuracy trade-off measured in Table~\ref{table:partition_delta}.

\noindent For a hidden dimension of $768$, an unpartitioned destination index requires $\lceil\log_2 768\rceil=10$ bits, giving $14$ bits per encoded weight ($4$ delta bits plus $10$ index bits). Partitioning allows the block identifier to be managed by the controller rather than repeated with every encoded weight. The per-weight index then identifies only the position within a block: $6$ bits for a 64-element block, $5$ for a 32-element block, and $4$ for a 16-element block. Including the 4-bit delta code gives total widths of $10$, $9$, and $8$ bits per weight, respectively.

\noindent Relative to the 8-bit quantized baseline, the 64-element scheme costs $2$ extra bits per weight (a $25\%$ storage overhead) in exchange for eliminating multipliers in the linear layers. The 16-element scheme has the same per-weight width as the 8-bit baseline.

\noindent When one encoded weight is supplied per cycle, these widths correspond to $10$, $9$, and $8$ bits/cycle of encoded-weight bandwidth for 64-, 32-, and 16-element partitions. Activation, output, and block-control traffic are accounted for separately and do not alter the per-weight encoding widths above.

\noindent As a direction for future work, an input redundancy mechanism could be incorporated to further reduce computation by skipping operations for repetitive video frames or regions within a frame. However, detecting similar frames or pixels requires input comparisons, which introduce additional overhead. To mitigate this cost, future work could explore integrating the architecture with front-end image-enhancement and super-resolution modules~\cite{parham, super} to extract redundancy information in advance, enabling input reuse and allowing redundant computations to be skipped more efficiently.

\section{Experimental Results}
\label{sec:experimental}

\subsection{Experimental Setup}
We evaluate DeVIT on the four architectures summarized in Table~\ref{table:transformers_benchmark}: ViT-B/16, DeiT, and Swin for image classification, and DETR-ResNet-50 for object detection. We use pretrained implementations from the Hugging Face and PyTorch model libraries.

Initially, we quantized the weights of these pretrained models to 8 bits. Next, we implemented DeVIT by sorting the weights, calculating the differences between consecutive weights, and approximating these differences as powers of two. Following offline modifications to the weights, we undertook RTL design for DeVIT with an emphasis on linear transformations and synthesized the design using the Synopsys Design Compiler within a 15\,nm technology process.

Partitioning reduces the partial-product buffer capacity relative to the unpartitioned design. We report the resulting accuracy, normalized computation load, and single-GEMM energy separately below.

\begin{table}[!t]
\centering
\caption{Properties of the evaluated transformer benchmarks.}
\label{table:transformers_benchmark}
\scriptsize
\setlength{\tabcolsep}{1.7pt}
\begin{tabular}{@{}lcccccc@{}}
\toprule
\textbf{Model} & \textbf{Embed.} & \textbf{Heads} & \textbf{Tokens} & \textbf{FFN} & \textbf{Params.} & \textbf{Layers} \\
\midrule
ViT~\cite{dosovitskiy2020image} & 768  & 12 & 577 & 3072 & 86  & 12 \\
DeiT~\cite{touvron2021training}& 768  & 12 & 578 & 3072 & 86  & 12 \\
Swin~\cite{Swin} & 1024 & 32 & 49  & 3072 & 88  & 24 \\
DETR~\cite{carion2020end} & 256  & 8  & 850 & 2048 & 41  & 6 \\
\bottomrule

\end{tabular}
\end{table}

\subsection{Results}

Figures~\ref{fig:delta-distribution-partitioning} and~\ref{fig:delta-distribution-FC} show that smaller partitions reduce the concentration of zero and one deltas, trading computation reuse for smaller buffers and indices. Table~\ref{table:partition_delta} reports accuracy for the 64- and 32-element configurations; the 16-element distributions are included to show the continuation of the partition-size trend. Relative to FP32, most reported score changes are below one percentage point. The largest changes are $3.11$ percentage points for Swin Top-1 and $2.53$ mAP points for DETR in the 32-element configuration, so we do not characterize every case as a sub-$2\%$ change.

The same partitioned, input-stationary method applies to both attention projections and the FFN. Figure~\ref{fig:delta-distribution-FC} reports the corresponding distributions for the two FFN matrices.

\begin{table}[!t]
\centering
\caption{Accuracy of partitioned DeVIT configurations.}
\label{table:partition_delta}
\scriptsize
\setlength{\tabcolsep}{2.5pt}
\begin{tabular}{@{}llcccc@{}}
\toprule
\textbf{Model} & \textbf{Metric} & \textbf{FP32} & \textbf{$\Delta$-64} & \textbf{$\Delta$-32} & \textbf{INT8} \\
\midrule
\textbf{ViT-B-16}~\cite{dosovitskiy2020image} & Top-1 & 83.71 & 83.48 & 82.82 & 81.60 \\
& Top-5 & 96.81 & 96.76 & 96.74 & 96.33 \\
\midrule
\textbf{DeiT}~\cite{touvron2021training} & Top-1 & 80.35 & 79.94 & 79.79 & 80.04 \\
& Top-5 & 93.91 & 93.70 & 93.37 & 93.93 \\
\midrule
\textbf{Swin}~\cite{Swin} & Top-1 & 82.92 & 80.04 & 79.81 & 83.15 \\
& Top-5 & 95.82 & 95.49 & 95.28 & 95.95 \\
\midrule
\textbf{DETR-Res50}~\cite{carion2020end} & mAP & 43.16 & 41.02 & 40.63 & 42.87 \\
\bottomrule
\end{tabular}
\end{table}

\subsection{Comparison with FACT}

To assess computational efficiency, we compare our method with the approach in FACT~\cite{fact}, which employs three skipping strategies: bypassing computations in the attention matrix, the first fully connected layer (FC1), and the second fully connected layer (FC2).

\begin{figure}[!t]
    \centering
    \includegraphics[trim=5 5 5 5,clip,width=\columnwidth]{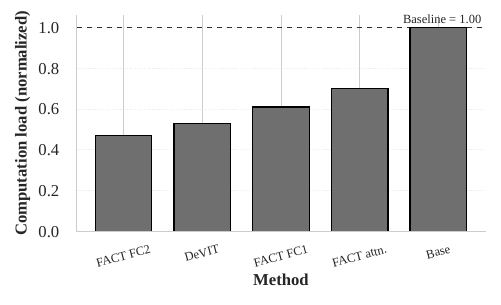}
    \caption{Normalized ViT computation load for DeVIT and the three FACT skipping strategies, relative to an unoptimized baseline ($1.00$).}
    \label{fig:vit-normalized-computation}
\end{figure}

Figure~\ref{fig:vit-normalized-computation} compares the corresponding normalized computation loads. DeVIT requires $0.53$ of the baseline computation, below FACT's FC1-skip ($0.61$) and attention-skip ($0.70$) operating points but above its most aggressive FC2-skip point ($0.47$). Relative to the attention-skip point, the reduction is approximately $24\%$. DeVIT obtains this reduction without a per-input prediction or out-of-order scheduling step.




\subsection{Energy Results}

\begin{figure}[!t]
    \centering
    \includegraphics[width=\columnwidth]{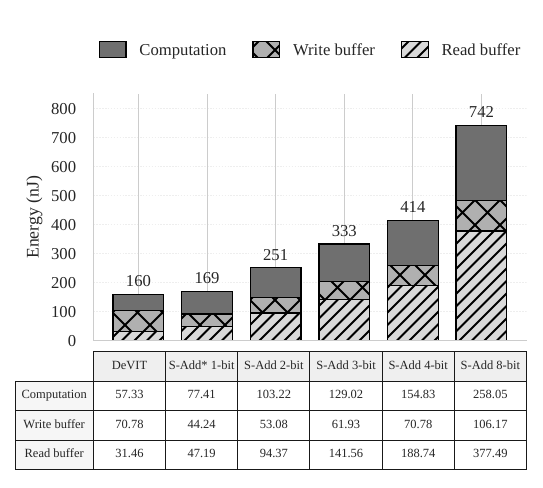}
    \caption{Single-GEMM energy breakdown for DeVIT and ShiftAddLLM (S-Add) configurations~\cite{you2024shiftaddllm}. Totals are printed above the bars; component values are listed in the embedded table.}
    \label{fig:energy-comparison}
\end{figure}

Figure~\ref{fig:energy-comparison} reports the evaluated single-GEMM energy. DeVIT totals $159.57$\,nJ: $57.33$\,nJ for computation, $70.78$\,nJ for buffer writes, and $31.46$\,nJ for buffer reads. The lowest-energy ShiftAddLLM configuration totals $168.84$\,nJ, making DeVIT approximately $5.5\%$ lower. These are per-configuration measurements, not a model-level speedup.

\section{Conclusion}

DeVIT replaces weight multiplications in dense transformer layers with reuse and shift--add operations over delta-encoded weights. Partition size controls the trade-off among reuse, index width, buffer capacity, and accuracy. DeVIT reaches $0.53$ normalized computation load and $159.57$\,nJ single-GEMM energy, $5.5\%$ below the lowest-energy ShiftAddLLM configuration shown. The worst measured accuracy changes---$3.11$ percentage points for Swin Top-1 and $2.53$ mAP points for DETR---define the current approximation scheme's practical limit.

\bibliographystyle{IEEEtran}
\bibliography{Main}
\end{document}